\PassOptionsToPackage{unicode}{hyperref}
\PassOptionsToPackage{hyphens}{url}
\documentclass[
]{article}
\usepackage{xcolor}
\usepackage[margin=1in]{geometry}
\usepackage{amsmath,amssymb}
\usepackage{iftex}
\ifPDFTeX
  \usepackage[T1]{fontenc}
  \usepackage[utf8]{inputenc}
  \usepackage{textcomp} 
\else 
  \usepackage{unicode-math} 
  \defaultfontfeatures{Scale=MatchLowercase}
  \defaultfontfeatures[\rmfamily]{Ligatures=TeX,Scale=1}
\fi
\usepackage{lmodern}
\ifPDFTeX\else
\fi
\IfFileExists{upquote.sty}{\usepackage{upquote}}{}
\IfFileExists{microtype.sty}{
  \usepackage[]{microtype}
  \UseMicrotypeSet[protrusion]{basicmath} 
}{}
\makeatletter
\@ifundefined{KOMAClassName}{
  \IfFileExists{parskip.sty}{%
    \usepackage{parskip}
  }{
    \setlength{\parindent}{0pt}
    \setlength{\parskip}{6pt plus 2pt minus 1pt}}
}{
  \KOMAoptions{parskip=half}}
\makeatother
\usepackage{color}
\usepackage{fancyvrb}

\DefineVerbatimEnvironment{Highlighting}{Verbatim}{commandchars=\\\{\}}
\newenvironment{Shaded}{}{}

\newcommand{\BuiltInTok}[1]{\textcolor[rgb]{0.00,0.50,0.00}{#1}}

\newcommand{\CommentTok}[1]{\textcolor[rgb]{0.38,0.63,0.69}{\textit{#1}}}

\newcommand{\KeywordTok}[1]{\textcolor[rgb]{0.00,0.44,0.13}{\textbf{#1}}}
\newcommand{\NormalTok}[1]{#1}
\newcommand{\OperatorTok}[1]{\textcolor[rgb]{0.40,0.40,0.40}{#1}}

\newcommand{\StringTok}[1]{\textcolor[rgb]{0.25,0.44,0.63}{#1}}
\newcommand{\VariableTok}[1]{\textcolor[rgb]{0.10,0.09,0.49}{#1}}

\usepackage{longtable,booktabs,array}
\usepackage{calc} 
\usepackage{etoolbox}
\makeatletter
\patchcmd\longtable{\par}{\if@noskipsec\mbox{}\fi\par}{}{}
\makeatother
\IfFileExists{footnotehyper.sty}{\usepackage{footnotehyper}}{\usepackage{footnote}}
\makesavenoteenv{longtable}
\usepackage{graphicx}
\makeatletter
\newsavebox\pandoc@box
\newcommand*\pandocbounded[1]{
  \sbox\pandoc@box{#1}%
  \Gscale@div\@tempa{\textheight}{\dimexpr\ht\pandoc@box+\dp\pandoc@box\relax}%
  \Gscale@div\@tempb{\linewidth}{\wd\pandoc@box}%
  \ifdim\@tempb\p@<\@tempa\p@\let\@tempa\@tempb\fi
  \ifdim\@tempa\p@<\p@\scalebox{\@tempa}{\usebox\pandoc@box}%
  \else\usebox{\pandoc@box}%
  \fi%
}
\def\fps@figure{htbp}
\makeatother
\providecommand{\tightlist}{%
  \setlength{\itemsep}{0pt}\setlength{\parskip}{0pt}}
\usepackage{bookmark}
\IfFileExists{xurl.sty}{\usepackage{xurl}}{} 
\hypersetup{
  hidelinks,
  pdfcreator={LaTeX via pandoc}}

\author{}
\date{}

\begin{document}
\title{ResearchQA: Benchmarking Citation-Grounded Question-Answering on Scientific Papers}
\author{Saba Imran \quad\quad Debanjum Singh Solanky \\\\ \textit{Khoj Inc.}}
\date{May 4, 2026}
\maketitle
\setlength{\parindent}{0pt}
\setlength{\parskip}{0.8em}

\begin{abstract}
Large language models are increasingly used to assist scientific
reading, but existing evaluation methods often fail to detect whether
answers are supported by verifiable citations. We introduce ResearchQA,
a benchmark of 6,211 single-paper question-answer pairs from 494
open-access papers spanning eight domains and four question types:
lookup, comprehension, multi-hop, and adversarial. ResearchQA is
designed for citation-grounded evaluation: it permits multiple valid
supporting passages for a claim and rewards grounded refusal when the
source paper does not support an answer. We evaluate eight leading
closed- and open-weight models in a citation-grounded chat-with-paper
setting using a deterministic citation matcher and an LLM-based rubric
evaluator. Citation-based metrics separate systems more clearly than
LLM-evaluator scores: section coverage and citation accuracy vary
substantially across models, while evaluator scores remain tightly
compressed. We further find that open-weight models approach the best
closed-model citation accuracy while achieving 3 to 6 times lower
per-example latency. We release the
\href{https://huggingface.co/datasets/khoj-ai/ResearchQA}{benchmark},
\href{https://github.com/khoj-ai/openpaper/tree/92cd89c85dbabb9bff8baa281e0678472ccb2ab4/server/evals}{evaluation
harness}, and evaluator prompt.

\end{abstract}
\subsection{Introduction}\label{introduction}

Researchers increasingly rely on large language models to read and
reason over scientific literature, where two properties matter more than
fluency: (1) every claim should be traceable to verifiable evidence in
the source, and (2) the answer should surface the complete, relevant
context rather than a confident subset of it. Fabricated or
misattributed citations propagate false assumptions and erode trust in
the findings that build on them. Standard evaluation methods fail to
measure these attributes: LLM evaluator scores reward plausible prose,
and embedding similarity passes fabricated-but-on-topic text. We
introduce ResearchQA, a benchmark that measures citation-grounded
answering directly, pairing LLM-generated questions with a deterministic
check that a cited passage actually appears in the paper it claims to.

Information retrieval benchmarks have a long history, but the existing
experiments did not satisfy our research constraints. They either target
a narrower corpus, the different granularity of evidence, or omit the
failure modes that matter most for a citation-grounded product.
\href{https://hotpotqa.github.io/}{\textbf{HotPotQA}} is grounded in
Wikipedia and constructed for multi-hop reasoning; it is well-curated
and widely used, but Wikipedia articles lack the dense numerical and
methodological content --- statistics, tables, methods sections,
findings --- that a ``chat with this paper'' feature is constantly asked
to ground. We eliminated it for that reason.

The closest prior art to our experiment is
\href{https://arxiv.org/abs/2105.03011}{\textbf{QASPER}}: a curated
corpus of NLP papers with questions, answers, and supporting evidence
extracted by human annotators. QASPER's design choices --- paper-level
scope, evidence-anchored answers, multiple question types --- directly
inform ours. The two reasons it was not enough on its own are that it
covers only NLP papers (we need a domain-diverse corpus to measure a
general-purpose research dataset), and that its annotation pipeline used
paid graduate-student labelers, which conflicted with the scale of the
rows-per-paper density we wanted. The open question that motivates a
substantial part of this work is whether a strong frontier LLM, paired
with a deterministic verification step against the source text, can
stand in for human annotators well enough to extend QASPER's design to a
multi-domain dataset at low cost.

Two methodology pitfalls shape the design that follows. LLM evaluator
scoring suffers from \textbf{rubric collapse} (the evaluators
concentrate at the endpoints of a 1--5 scale unless every level is
anchored); citation evaluation suffers from a verbatim-vs-paraphrase
tradeoff between substring matching (which fails on PDF extraction
noise) and embedding similarity (which passes fabricated-but-on-topic
quotes). \emph{Methods} describes how we mitigate rubric collapse and
the verbatim-vs-paraphrase tradeoff; \textbf{same-family bias} (an
evaluator over-rates responses from its own family during grading) is a
known confound we characterize but do not eliminate in v1, discussed in
\emph{Limitations}.

Three measurement needs follow: (1) a metric that fails fabricated
citations independently of answer quality, so models that prose-glue
plausible numbers do not score identically to models that quote the
paper; (2) adversarial robustness as a first-class score, rewarding
refusal-with-grounded-evidence rather than burying it in an aggregate;
(3) multi-hop reasoning measured by \emph{which sections are
integrated}, not just whether the prose mentions multiple things.

\subsection{Methods}\label{methods}

The benchmark has three concrete artifacts --- the dataset, the harness
that runs models against it, and the grader that scores their outputs
--- plus two pieces of grader machinery (the citation matcher and the
LLM evaluator) that account for most of the signal in the results. We
describe each in turn.

\subsubsection{Dataset construction}\label{dataset-construction}

\pandocbounded{\includegraphics[keepaspectratio,alt={Dataset Construction Pipeline}]{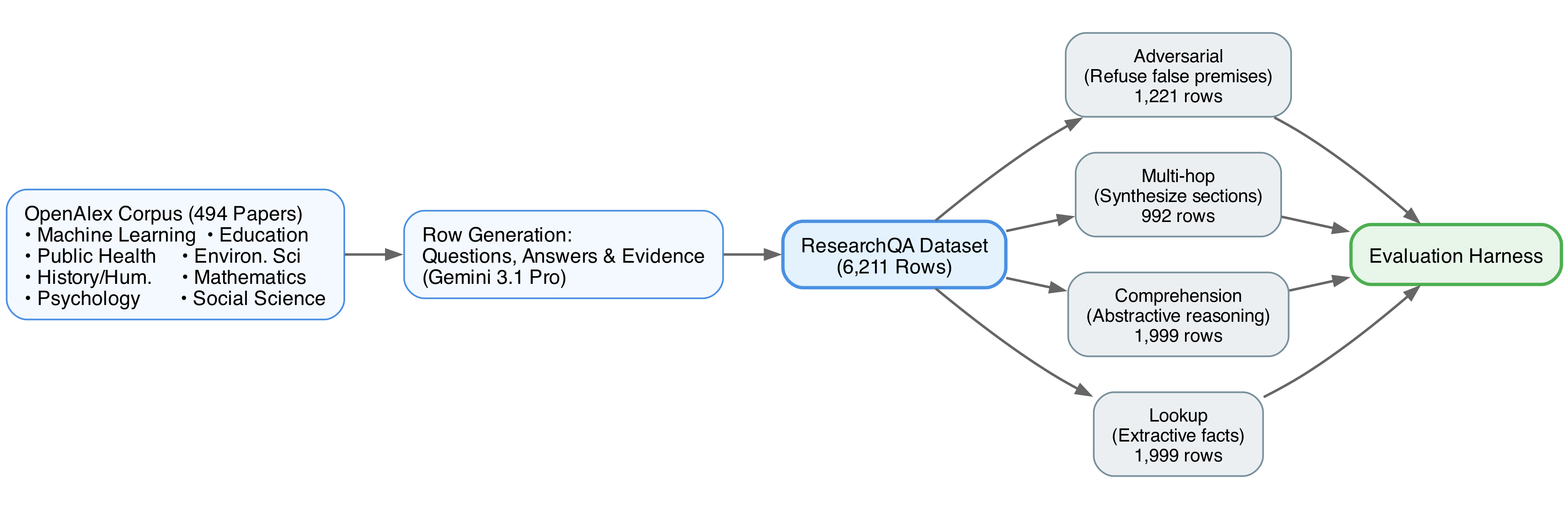}}
\emph{Figure 1: ResearchQA dataset construction pipeline. The corpus is
drawn from OpenAlex across 8 domains. Questions and expected citations
are generated by Gemini 3.1 Pro before benchmarking via the evaluation
harness.}

The paper corpus is sourced from
\href{https://openalex.org/}{OpenAlex}'s open-access subset across eight
domains (machine learning, public health, education, environmental
science, history and humanities, mathematics, psychology, and social
science), selected to balance technical density (numbers, methods,
tables) with prose-heavy disciplines that exercise abstractive
comprehension. At the time of this paper, the benchmark contains 494
papers; both the PDF (for the harness's input pipeline) and the
extracted full text (for grading-time citation matching) are stored.

Question/answer/evidence triples are generated per paper chunk by Gemini
3.1 Pro under a structured-output schema that constrains the model's
output shape and validates each row against the schema before it is
written.

\pandocbounded{\includegraphics[keepaspectratio,alt={Row Generation Fan-out}]{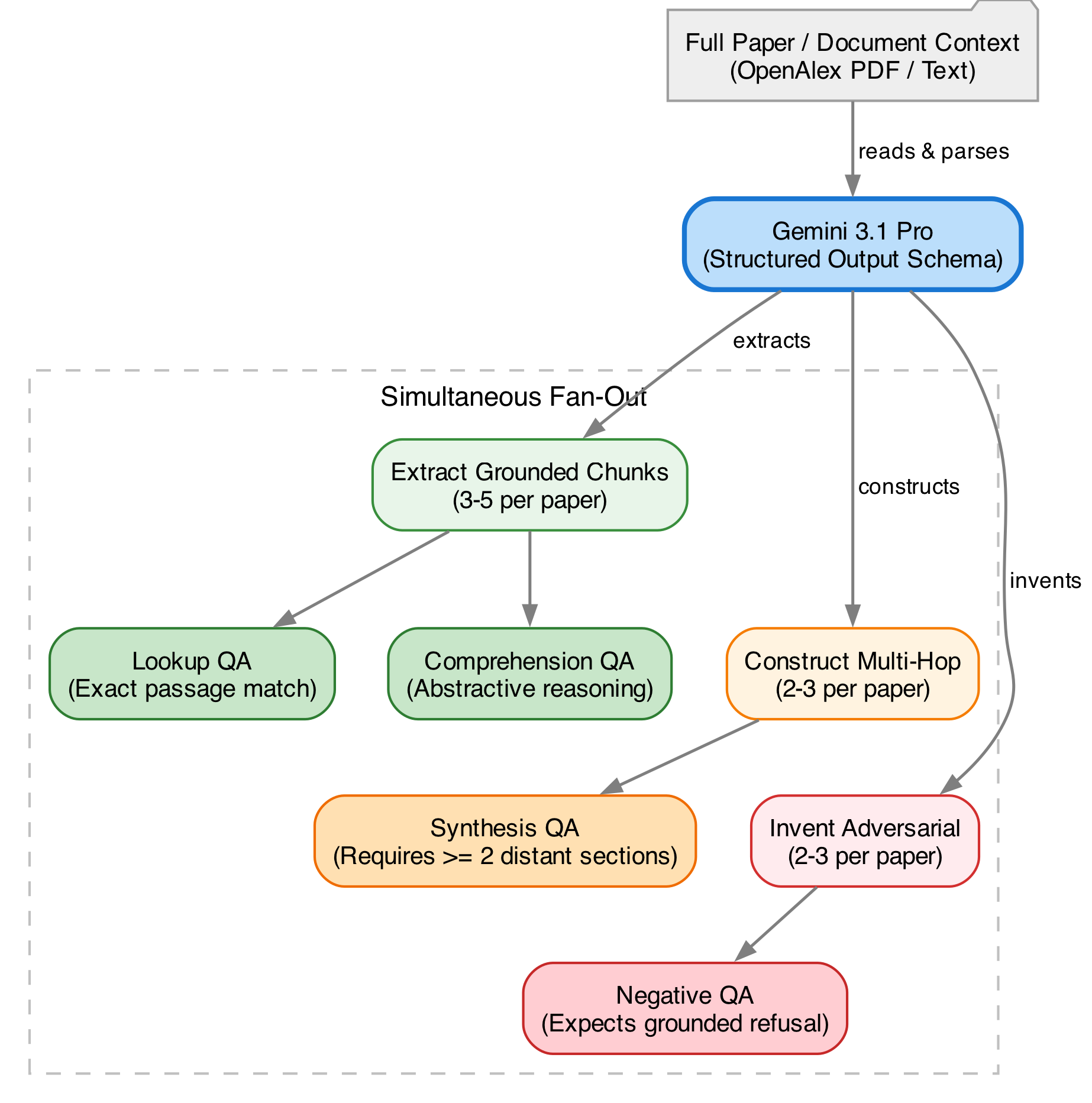}}
\emph{Figure 2: Row generation fan-out. The agent reads the paper and
simultaneously extracts grounded chunks (pairing them with lookup and
comprehension questions), constructs multi-hop questions requiring
cross-section synthesis, and generates adversarial questions containing
false premises.}

At the time of writing, the dataset has 6,211 rows with a four-way
taxonomy. The four question types are:

\begin{itemize}
\tightlist
\item
  \textbf{\texttt{lookup}} (extractive): a factual question whose answer
  is a single passage in the paper --- e.g.~``what was the sample
  size?'' --- testing whether the model can locate and quote the right
  span.
\item
  \textbf{\texttt{comprehension}} (abstractive): an open-ended question
  about themes, methodology, or implications that requires the model to
  synthesize a passage rather than copy it --- e.g.~``what is the
  authors' critique of prior work?''.
\item
  \textbf{\texttt{multi\_hop}}: a question whose answer cannot be
  obtained from any single passage and requires combining evidence from
  two or more distant sections --- e.g.~comparing a result against a
  stated baseline, or computing a derived quantity from numbers spread
  across methods, results, and tables.
\item
  \textbf{\texttt{adversarial}}: a question containing a false premise
  or asking about something the paper does not address --- e.g.~asking
  about a placebo arm in a single-arm study --- where the correct
  behavior is to identify the false premise and refuse rather than
  fabricate an answer.
\end{itemize}

Multi-hop questions are the hardest to generate well, since the
LLM-author must construct a question whose answer cannot be obtained
from any single passage. The structured-output schema enforces this at
generation time (full schema in Appendix B): each multi-hop row carries
at least two \texttt{SectionEvidence} blocks pointing to distinct
sections, a \texttt{reasoning\_chain} naming the dependency between
hops, and a \texttt{judge\_rubric} criterion requiring integration
across sections. The most common patterns the generator produces are
comparing a result to a stated baseline, checking whether a discussion
caveat invalidates a headline claim, and computing a derived quantity
from numbers spread across methods, results, and tables.

Based on the evidence blocks, scoring uses \textbf{AND across sections,
OR within each section's alternatives}: a model satisfies a row when it
cites at least one alternative from every required section. For lookup
and comprehension rows this typically resolves to a single
\texttt{SectionEvidence} with multiple alternatives, so coverage is
binary. For multi-hop rows the list contains one
\texttt{SectionEvidence} per required section, so coverage is fractional
--- a model that addresses two of three required sections gets 0.667
credit.

\subsubsection{Benchmark harness}\label{benchmark-harness}

The harness routes each question through Open Paper's full
chat-with-paper pipeline --- retrieval over the indexed paper, the
citation contract that requires the model to anchor each claim to a
quoted passage, and the structured-output parser that extracts citations
from the model's response.

\subsubsection{Metrics}\label{metrics}

Each row is scored on up to six metrics, four deterministic and two
LLM-evaluated.

The four deterministic metrics are:

\begin{itemize}
\tightlist
\item
  \textbf{\texttt{citation\_precision}}: the fraction of the model's
  citations that match some alternative in some required section ---
  measuring how many of the model's citations were ``useful'' for the
  rubric.
\item
  \textbf{\texttt{section\_coverage}}: the fractional satisfaction of
  required sections, AND across sections and OR within each section's
  alternatives --- measuring whether the model touched every required
  section.
\item
  \textbf{\texttt{citation\_accuracy}}: the fraction of the model's
  citations that are verifiable substrings of the paper's raw text,
  after the normalization pass described below --- measuring whether the
  model fabricated quotes.
\item
  \textbf{\texttt{refusal\_correctness}} (adversarial rows only): 1.0 if
  the model refused entirely or every citation it produced is grounded
  in the paper, 0.0 if any cited passage is fabricated --- measuring
  whether the model invented evidence to support its refutation of a
  false premise.
\end{itemize}

The two LLM-evaluated metrics, each on a 1--5 scale with per-level
anchored definitions described below, are:

\begin{itemize}
\tightlist
\item
  \textbf{\texttt{factual\_accuracy}}: whether every factual claim in
  the answer is consistent with the expected answer.
\item
  \textbf{\texttt{completeness}}: whether every key point in the
  expected answer is addressed.
\end{itemize}

We deliberately measure the deterministic and evaluator metrics
independently so that the citation-grounding signal cannot be drowned
out by evaluator noise; this separation also makes the evaluator's
failure modes visible (when the deterministic metrics show wide variance
and the evaluator metrics saturate, we know to look at the evaluator).

\subsubsection{PDF-aware citation
matcher}\label{pdf-aware-citation-matcher}

The citation matcher decides whether a model's quoted citation appears
in the paper's extracted text. Naively this is a substring check, but
\href{https://pypi.org/project/PyPDF2/}{PyPDF2}'s text extraction
introduces enough noise --- line-break hyphenation, mid-word spaces,
ligatures, smart quotes --- that faithfully-quoted citations regularly
fail an exact match. The matcher absorbs this noise through a
normalization pipeline applied to both sides, with a final
whitespace-stripped fallback that reclaims citations broken by mid-word
PDF artifacts; full implementation details are in the released code. We
chose this over embedding similarity because the failure modes diverge:
a normalization-based matcher rejects fabricated content that does not
appear in the paper, while an embedding matcher would pass a
fabricated-but-on-topic quote that lives near real paper content in
embedding space.

\subsubsection{LLM Evaluator}\label{llm-evaluator}

The LLM evaluator scores \texttt{factual\_accuracy} and
\texttt{completeness} on a 1--5 scale, with each level anchored by
concrete criteria rather than only the endpoints.

For \texttt{factual\_accuracy}:

\begin{itemize}
\tightlist
\item
  \textbf{5} --- every factual claim in the actual answer is consistent
  with the expected answer with no fabricated specifics.
\item
  \textbf{4} --- exactly one minor inaccuracy (a misstated number close
  to correct, a slightly wrong attribution).
\item
  \textbf{3} --- one substantive error or two-to-three minor ones.
\item
  \textbf{2} --- multiple substantive errors, or one error that
  undermines the main claim.
\item
  \textbf{1} --- the core claim contradicts the expected answer, or the
  answer is fabricated wholesale.
\end{itemize}

For \texttt{completeness}:

\begin{itemize}
\tightlist
\item
  \textbf{5} --- every key point in the expected answer is addressed.
\item
  \textbf{4} --- exactly one minor omission.
\item
  \textbf{3} --- one substantive omission or two-to-three minor ones.
\item
  \textbf{2} --- multiple substantive omissions.
\item
  \textbf{1} --- the core point is not addressed at all.
\end{itemize}

Under the anchored rubric, \texttt{completeness} demonstrates meaningful
variance. This pattern is consistent with concurrent work showing that
locked, evidence-anchored rubrics reduce evaluator instability
(\href{https://arxiv.org/abs/2601.08654}{Hong et al., 2025}) and that
knowledge-grounded rubrics produce more discriminating evaluator
judgments than surface-heuristic ones
(\href{https://arxiv.org/abs/2603.11027}{Song et al., 2026}).
\texttt{factual\_accuracy} still saturated at 5.000 for Gemini 3.1 Pro
on this slice --- interpretable as ``the model is genuinely accurate on
the factual claims this dataset asks about'' rather than as residual
rubric collapse, since weaker models in the comparison did receive sub-5
scores. Currently a single Gemini evaluator grades all systems including
itself; same-family bias is a known limitation discussed in
\emph{Limitations \& Future Work}.

\pandocbounded{\includegraphics[keepaspectratio,alt={LLM Judge Architecture}]{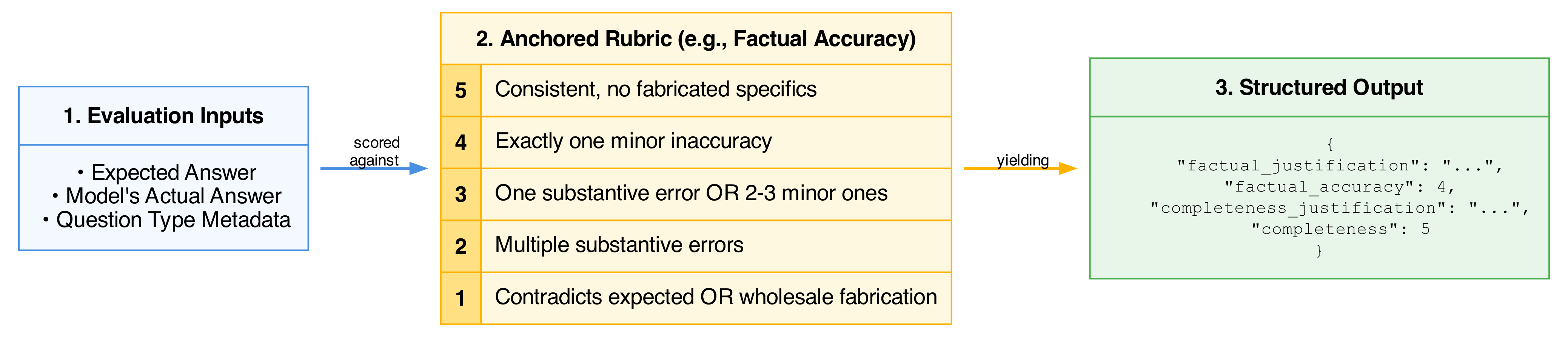}}
\emph{Figure 3: LLM evaluator flow. The evaluator operates on a strict
1--5 anchored rubric designed to combat score collapse, and it must
output a detailed justification before outputting the final numerical
score to enforce a chain-of-thought analysis.}

\subsection{Results}\label{results}

We ran the benchmark end-to-end through the Open Paper chat-with-paper
harness against eight LLMs in two configurations each --- a more
intelligent model and a faster model: \texttt{gemini-3.1-pro-preview} /
\texttt{gemini-3-flash-preview}, \texttt{gpt-5.4} / \texttt{gpt-4.1},
\texttt{claude-opus-4-7} / \texttt{claude-haiku-4-5}, and the
Cerebras-hosted \texttt{gpt-oss-120b} / \texttt{zai-glm-4.7}. Each
system answered the same 100-row evenly-spaced sample of the dataset,
and each row was graded by the deterministic citation metrics and the
LLM evaluator described in \emph{Methods}. Aggregate results are shown
below, ranked by a composite of citation-grounding and answer-quality
scores.

{\def\LTcaptype{none} 
\begin{longtable}[]{@{}
  >{\raggedright\arraybackslash}p{(\linewidth - 14\tabcolsep) * \real{0.2708}}
  >{\raggedleft\arraybackslash}p{(\linewidth - 14\tabcolsep) * \real{0.1250}}
  >{\raggedleft\arraybackslash}p{(\linewidth - 14\tabcolsep) * \real{0.1146}}
  >{\raggedleft\arraybackslash}p{(\linewidth - 14\tabcolsep) * \real{0.1146}}
  >{\raggedleft\arraybackslash}p{(\linewidth - 14\tabcolsep) * \real{0.0938}}
  >{\raggedleft\arraybackslash}p{(\linewidth - 14\tabcolsep) * \real{0.0938}}
  >{\raggedleft\arraybackslash}p{(\linewidth - 14\tabcolsep) * \real{0.0833}}
  >{\raggedleft\arraybackslash}p{(\linewidth - 14\tabcolsep) * \real{0.1042}}@{}}
\toprule\noalign{}
\begin{minipage}[b]{\linewidth}\raggedright
Model
\end{minipage} & \begin{minipage}[b]{\linewidth}\raggedleft
Cite. Prec
\end{minipage} & \begin{minipage}[b]{\linewidth}\raggedleft
Sect. Cov
\end{minipage} & \begin{minipage}[b]{\linewidth}\raggedleft
Cite. Acc
\end{minipage} & \begin{minipage}[b]{\linewidth}\raggedleft
Refusal
\end{minipage} & \begin{minipage}[b]{\linewidth}\raggedleft
Factual
\end{minipage} & \begin{minipage}[b]{\linewidth}\raggedleft
Compl.
\end{minipage} & \begin{minipage}[b]{\linewidth}\raggedleft
Lat. (s)
\end{minipage} \\
\midrule\noalign{}
\endhead
\bottomrule\noalign{}
\endlastfoot
gemini-3.1-pro-preview & 0.723 & \textbf{0.955} & 0.817 & \textbf{0.870}
& \textbf{5.000} & \textbf{4.940} & 18.3 \\
gpt-5.4 & 0.647 & 0.902 & 0.834 & 0.783 & 4.970 & 4.870 & 12.3 \\
zai-glm-4.7 & 0.744 & 0.776 & 0.840 & 0.783 & 4.906 & 4.875 & 5.6 \\
gpt-4.1 & \textbf{0.776} & 0.755 & 0.786 & \textbf{0.870} & 4.910 &
4.730 & 10.9 \\
claude-opus-4-7 & 0.569 & 0.876 & 0.809 & 0.739 & 4.980 & \textbf{4.980}
& 30.4 \\
claude-haiku-4-5 & 0.716 & 0.843 & \textbf{0.845} & 0.739 & 4.802 &
4.659 & 9.8 \\
gemini-3-flash-preview & 0.491 & 0.913 & 0.835 & 0.652 & 4.990 & 4.940 &
14.8 \\
gpt-oss-120b & 0.660 & 0.636 & 0.700 & 0.478 & 4.745 & 4.571 &
\textbf{2.9} \\
\end{longtable}
}

\emph{Table 1: Aggregate performance of all evaluated models across the
OpenPaper benchmark. Models are sorted by a composite of
citation-grounding and answer-quality scores. Best scores in each column
are bolded.}

\textbf{The citation-grounding metrics discriminate; the evaluator
metrics largely saturate.} The largest cross-provider spread sits in
\texttt{refusal\_correctness} (0.478--0.870, a 39-point gap on a 0--1
scale) and \texttt{section\_coverage} (0.636--0.955, a 32-point gap).
\texttt{citation\_precision} (0.491--0.776, a 28-point gap) and
\texttt{citation\_accuracy} (0.700--0.845, a 14-point gap) follow. The
LLM evaluator, by contrast, separates the field by far less:
\texttt{factual\_accuracy} (4.745--5.000, a 0.255-point gap on a 1--5
scale) and \texttt{completeness} (4.571--4.980, a 0.409-point gap). We
read this as ``current-generation models are uniformly correct on the
kinds of factual claims the dataset asks about, but they vary
substantially in \emph{which evidence they choose to cite and whether
that evidence is verifiable}'' --- exactly the distinction the citation
metrics were designed to expose.

\pandocbounded{\includegraphics[keepaspectratio,alt={Model Performance Profiles}]{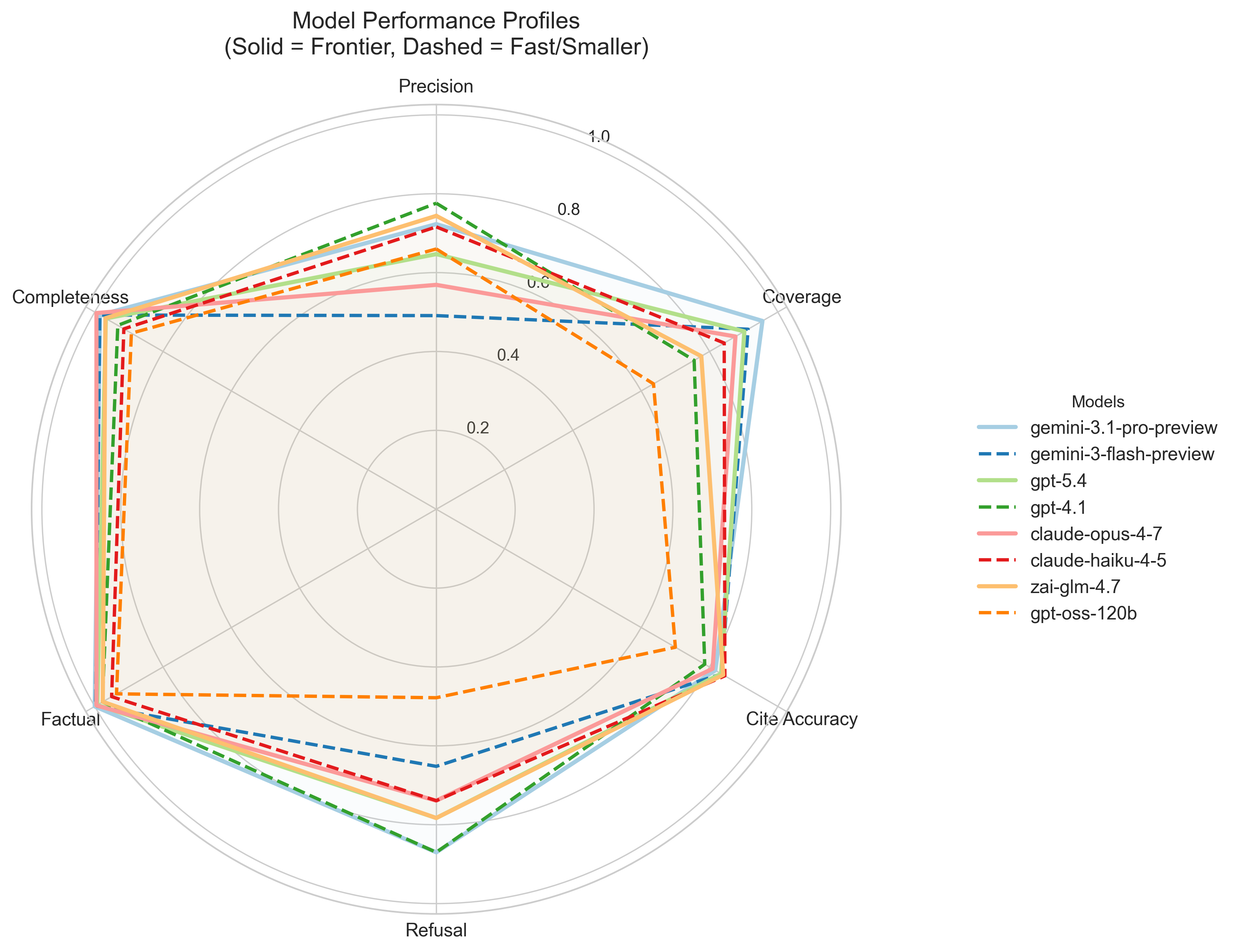}}
\emph{Figure 4: Model Performance Profiles. Frontier models (solid) and
their faster counterparts (dashed) plotted across normalized metrics.
Evaluator metrics (Factual, Completeness) are saturated at the
perimeter, while strict citation metrics (Precision, Coverage, Refusal)
reveal significant differences in model behavior. No single system
dominates all dimensions.}

\textbf{No single system dominates.} Gemini 3.1 Pro leads on composite
quality, but four different models top at least one metric in Table 1
--- every system is best at something and weakest at something else.
Claude Opus 4.7 illustrates the pattern: it ties for the most complete
answers but has the lowest citation precision, the signature of a
verbose model that finds the right answers and then over-cites. The two
Cerebras-hosted models stake out a separate speed-quality frontier: Zai
GLM 4.7 reaches near-frontier quality at roughly a third of Gemini 3.1
Pro's latency, and GPT-OSS-120B is six times faster than Gemini 3.1 Pro
at the cost of meaningful ground on refusal correctness and coverage.
The faster variants from the closed-model providers do not form a
comparable frontier --- they sit in the same latency band as the default
models.

\textbf{Latency varies by an order of magnitude.} End-to-end per-row
latency ranges from 2.9 s (GPT-OSS-120B) to 30.4 s (Claude Opus 4.7).
Six of the eight models run under 15 seconds per row; only the two
slowest reasoning models --- Gemini 3.1 Pro at 18.3 s and Claude Opus
4.7 at 30.4 s --- cross that threshold.

\pandocbounded{\includegraphics[keepaspectratio,alt={Speed-Quality Frontier}]{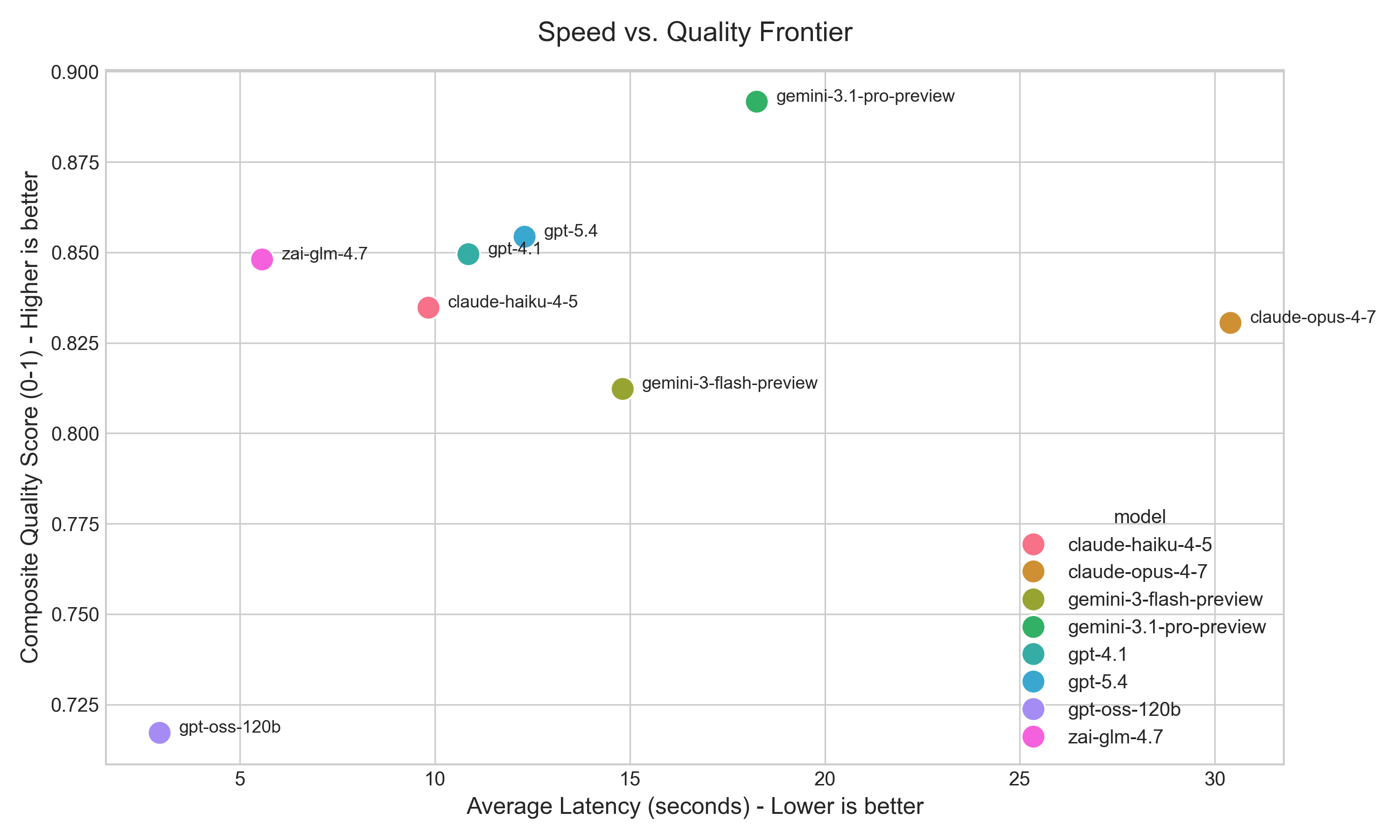}}
\emph{Figure 5: Speed-Quality Frontier. Average latency plotted against
a composite quality score (average of citation and normalized evaluator
metrics). Fast models like \texttt{gpt-oss-120b} and
\texttt{zai-glm-4.7} establish the low-latency edge, while
\texttt{gemini-3.1-pro-preview} leads on quality.}

\textbf{Multi-hop is the consistently hardest slice.} Across the eight
systems, multi-hop questions show the widest spread on every metric:
\texttt{section\_coverage} (0.542--0.958, a 42-point gap on a 0--1
scale), \texttt{citation\_accuracy} (0.444--0.821, a 38-point gap),
\texttt{factual\_accuracy} (4.417--5.000, a 0.583-point gap on a 1--5
scale), and \texttt{completeness} (4.000--5.000, a full 1.000-point
gap). Lookup and comprehension are partially saturated --- every system
hits at least 0.703 on lookup citation accuracy, and
\texttt{factual\_accuracy} on lookup never drops below 4.829 ---
confirming the original observation from the spec that single-passage
retrieval has become an easy slice for current models. The decision to
add multi-hop and adversarial categories after observing saturation on
the original lookup-only dataset is borne out by the spread numbers:
those two categories now drive most of the cross-system signal.

\pandocbounded{\includegraphics[keepaspectratio,alt={Multi-hop Performance Drop}]{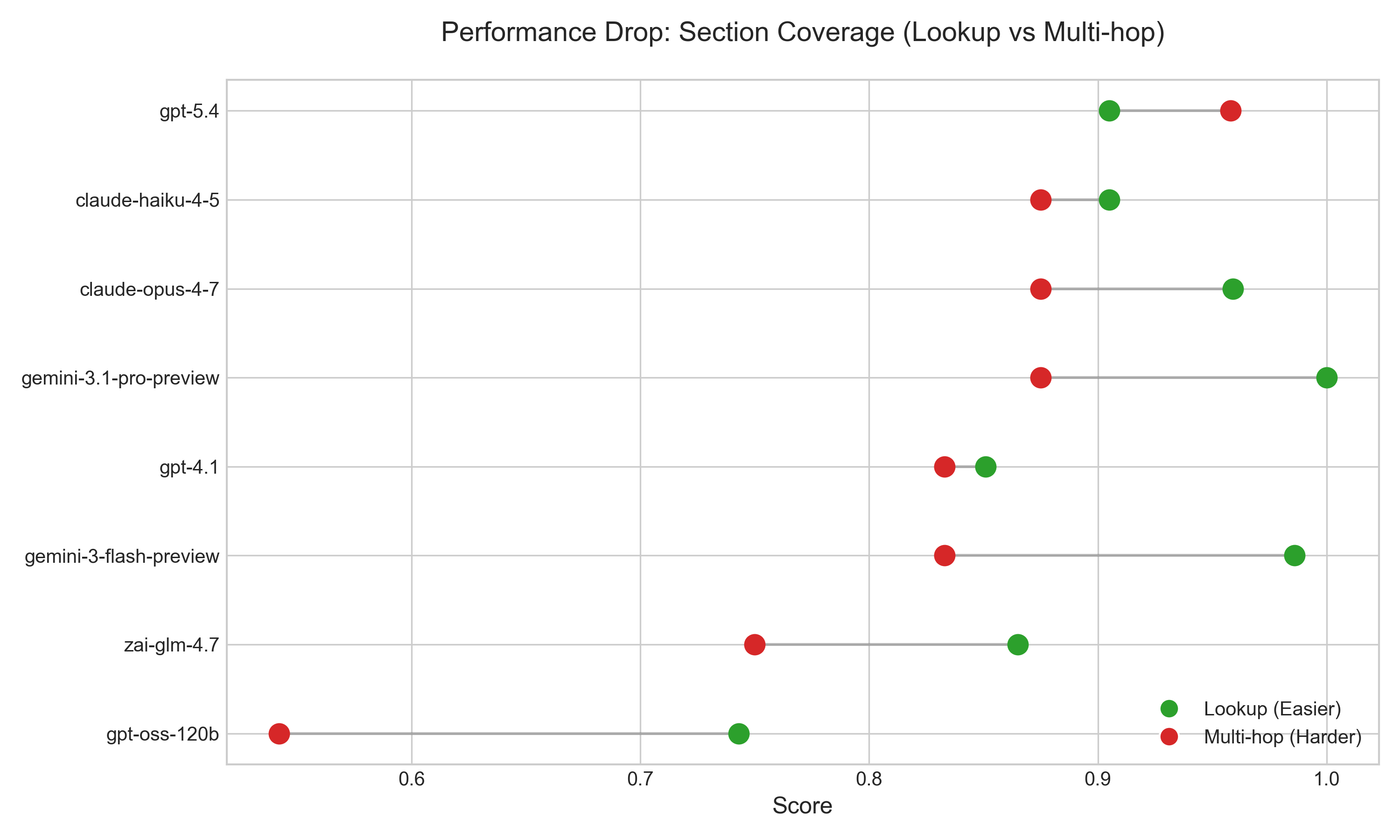}}
\emph{Figure 6: Multi-hop Performance Drop. A comparison of section
coverage on lookup versus multi-hop questions. While most models achieve
near-perfect coverage on simple lookups (green), performance degrades
and spreads significantly on multi-hop questions (red).}

\textbf{Adversarial behavior splits the field cleanly.} On adversarial
questions, the model is asked something whose premise is false or whose
answer is not in the paper. The correct response is to identify the
false premise and either refuse or refute with grounded evidence. The
new \texttt{refusal\_correctness} metric --- which marks a row correct
when every cited passage is verifiable in the paper text (i.e., the
model didn't fabricate evidence to support its refutation) --- produces
the largest cross-provider gap of any metric: Gemini 3.1 Pro and GPT-4.1
score 0.870, the next tier (GPT-5.4, Zai GLM 4.7) scores 0.783, the
Claude pair scores 0.739, Gemini 3 Flash scores 0.652, and GPT-OSS-120B
scores 0.478. The evaluator's \texttt{factual\_accuracy} on adversarial
spans 4.478--5.000, a similar shape to refusal but compressed into the
top of the scale; the deterministic refusal metric gives a more readable
signal.

\textbf{Performance varies significantly by subject matter.} Across the
100-question sample, difficulty is not uniform. Aggregating the strict
citation metrics (Precision, Coverage, Accuracy, Refusal) reveals a
clear gradient of domain difficulty. Models perform relatively well and
consistently on topics like \emph{Economics} and
\emph{History/Humanities}. Conversely, \emph{Education} and
\emph{Machine Learning} proved to be the most challenging domains,
exhibiting the lowest median scores and the widest variance across
models. Intermediate domains like \emph{Psychology} maintain a
relatively strong median, while \emph{Biology} shows a notable spread of
lower-performing outliers despite a decent median.

\pandocbounded{\includegraphics[keepaspectratio,alt={Domain Difficulty Variation}]{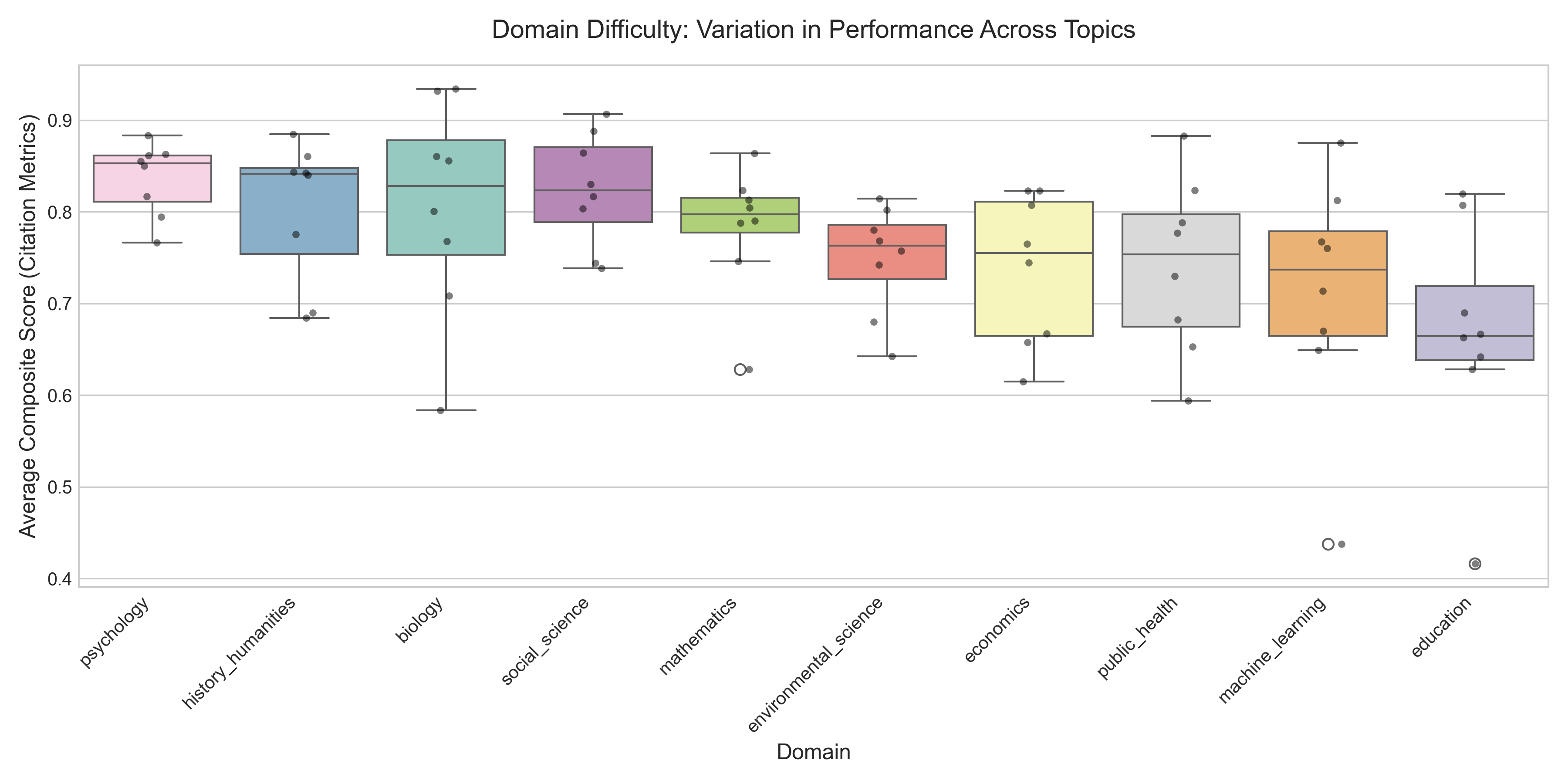}}
\emph{Figure 7: Domain Difficulty Variation. The domains ordered from
highest to lowest median performance on the strict citation metrics.
Black dots represent individual model scores, illustrating the lower
medians and widening spreads on domains like Machine Learning and
Education.}

\textbf{Implications for the harness and the benchmark.} The
deterministic citation metrics are doing the work the LLM evaluator
cannot: they flag when a model fabricates a citation, when it ignores a
required section in a multi-hop question, and when it confidently rebuts
a false premise with invented evidence. The evaluator metrics are not
useless --- \texttt{completeness} does discriminate models, and
\texttt{factual\_accuracy} would presumably discriminate weaker base
models if we added them --- but in the current frontier-tier comparison
the citation metrics carry most of the signal. We observe that splitting
measurements along citation-grounding helps measure the differentiator
directly.

\subsection{Related Work}\label{related-work}

ResearchQA sits at the intersection of three lines of prior work:
scientific paper question-answering, citation-grounded language
modeling, and the broader move toward benchmarks that measure end-to-end
practical task performance. We summarize each line here and locate our
contribution within it.

\textbf{Scientific paper QA.} The most direct ancestor is QASPER
(\href{https://aclanthology.org/2021.naacl-main.365/}{Dasigi et al.,
2021}), a 5,049-question dataset over NLP papers with human-annotated
supporting evidence and a four-way question taxonomy whose design
directly informs ours; ResearchQA differs in scope (eight domains vs NLP
only), annotation pipeline (LLM-generated and matcher-verified vs
human-labeled), and metric set (we add deterministic citation matching
and an adversarial-refusal slice). Two adjacent benchmarks were
considered but rule themselves out: ELI5
(\href{https://aclanthology.org/P19-1346/}{Fan et al., 2019}) targets
free-form long-form QA without citation grounding, and ScienceQA
(\href{https://proceedings.neurips.cc/paper_files/paper/2022/hash/11332b6b6cf4485b84afadb1352d3a9a-Abstract-Conference.html}{Lu
et al., 2022}) tests multimodal multiple-choice reasoning over K--12
science topics rather than free-form citation-grounded QA over research
papers.

\textbf{Multi-hop and long-document QA.} Reasoning over scientific
papers is closer to the long-document QA literature than to web-style
multi-hop benchmarks. \href{https://aclanthology.org/2024.tacl-1.9/}{Liu
et al.~(2024)} characterize the ``lost in the middle'' effect ---
language models systematically lose information from the middle of long
contexts --- which is the failure mode our \texttt{multi\_hop} slice is
designed to surface and our \texttt{section\_coverage} metric penalizes.
Adjacent work in the area includes long-context multi-document QA
benchmarks (\href{https://aclanthology.org/2024.emnlp-main.322/}{Wang et
al., 2024a}), discourse-structure-aware retrieval
(\href{https://aclanthology.org/2023.findings-emnlp.972/}{Nair et al.,
2023}), and iterative or graph-based retrieval methods
(\href{https://doi.org/10.1145/3701716.3716889}{Jiang et al., 2025};
\href{https://doi.org/10.1609/aaai.v38i17.29889}{Wang et al., 2024b})
--- candidate strategies for the harness's retrieval layer rather than
benchmarks we compare against.

\textbf{Citation-grounded answering.} The most direct prior work on the
behavior our benchmark measures is GopherCite
(\href{https://arxiv.org/abs/2203.11147}{Menick et al., 2022}), which
trains a language model to support its answers with verified verbatim
quotes. The metric we call \texttt{citation\_accuracy} --- does the
cited string actually appear in the source --- is exactly the
verification step GopherCite trains models to perform; our contribution
is not a new training method but a measurement infrastructure that
grades any model's citation behavior, including models not trained
explicitly for it. The verbatim-vs-paraphrase tradeoff we identify in
\emph{Introduction} is also discussed in the GopherCite work, which
lands on the same side we do: verbatim is the right contract for
groundedness even though it sometimes punishes faithful paraphrase.

\textbf{Benchmarks for real-world task performance.} ResearchQA fits a
recent trend toward benchmarks grounded in economically or
intellectually meaningful tasks: GDPval
(\href{https://arxiv.org/abs/2510.04374}{Patwardhan et al., 2025}) on
real work products across 44 occupations, APEX
(\href{https://arxiv.org/abs/2509.25721}{Vidgen et al., 2025}) on
professional-task performance in banking, consulting, law, and primary
care, the Remote Labor Index
(\href{https://arxiv.org/abs/2510.26787}{Mazeika et al., 2025}) on
end-to-end agent performance on remote-work projects, and SWE-FFICIENCY
(\href{https://arxiv.org/abs/2511.06090}{Ma et al., 2025}) on
repository-level optimization. ResearchQA is the equivalent for reading,
citing, and reasoning over research papers; the citation-grounding focus
reflects what makes that labor worth automating in the first place --- a
researcher's time is best spent reading evidence, not verifying the
model invented it.

\subsection{Limitations \& Future Work}\label{limitations-future-work}

The most important limitation of v1 is the absence of human-annotated
ground truth. The dataset's questions, expected answers, and evidence
passages are LLM-generated and verified only by the citation matcher and
the LLM evaluator during benchmarking; our team has spot-checked
individual rows during development, but no stratified subset has been
independently validated. We would like to recruit domain-expert labelers
--- graduate students or postdocs with literature-review experience in
each represented field --- to validate a stratified sample, report
inter-annotator agreement, and calibrate the LLM evaluator against human
scores. Until that work happens, every metric in this paper rests on the
assumption that Gemini 3.1 Pro's question generation and our matcher's
verdicts agree with what an expert human reader would say; we believe
this assumption holds for most rows but cannot prove it.

All LLM-evaluator scores reported here come from a single Gemini model
grading every system, including itself. Same-family bias is a documented
effect, and the close-to-ceiling factual-accuracy scores for the
Gemini-family systems in \emph{Results} should be read with that in
mind. A future release will use either a heterogeneous evaluator
(Deepseek grading Gemini, GPT grading Claude, and so on) or a
multi-evaluator consensus to remove this confound.

The same team that designed the metrics, wrote the matcher, and tuned
the rubric also generated and curated the dataset. This collapses
several roles that in larger benchmark efforts are kept separate, and
choices that improve our reported scores are difficult to distinguish
from choices that improve measurement quality. The \emph{Construction
tradeoffs} and \emph{Appendix A: Alternatives considered} sections
record the decisions most likely to be contested; readers should treat
the metric/dataset co-design as a known confound.

Every question in the dataset was authored by Gemini 3.1 Pro under a
structured-output schema. The dataset's question style, difficulty
distribution, and category boundaries therefore reflect one model
family's choices about what counts as a ``lookup'' or ``comprehension''
or ``multi-hop'' question. Models from the same family may have a small
advantage on this distribution that does not generalize to questions
written by humans or by a different LLM family. Diversifying the
generator across model families is straightforward and worth doing in a
future release.

The corpus and all questions are English-language. ResearchQA's results
should not be read as evidence about model behavior on non-English
scientific literature; that gap remains an open problem in the broader
benchmarking landscape and one this paper does not address.

The corpus is drawn from OpenAlex's open-access subset across eight
domains, which is not a representative sample of all scientific
literature. Closed-access papers and venues are absent. Within the
included domains, our 494-paper sample was further filtered for PDF
availability and successful text extraction, which biases against papers
with image-heavy layouts, complex tables, or scanned originals ---
exactly the papers a citation-grounded reader is most useful for.

The dataset is publicly released on HuggingFace
(\href{https://huggingface.co/datasets/khoj-ai/ResearchQA}{\texttt{khoj-ai/ResearchQA}}),
which means future model versions may have seen these papers and
questions during training. Benchmarks suffer from contamination over
time, and ResearchQA is no exception. We accepted this as the cost of an
open benchmark; readers comparing model versions released after the
dataset's publication date should treat their scores with appropriate
skepticism.

ResearchQA measures text-grounded understanding. The harness feeds the
native PDF to models that can read it directly and falls back to
extracted full text only for models that cannot, so multimodal-capable
systems do exercise their visual capabilities on the input side. The
benchmark itself, however, is text-centric: questions are authored from
the paper's text and never require interpreting a figure, chart,
diagram, or rendered table to answer, and the citation matcher grades
against extracted text. A natural and valuable extension is a benchmark
built specifically for multimodal scientific understanding --- questions
that target visual content directly, with grounding and grading that
verify a model's answer against figures and diagrams rather than prose
alone.

The single-paper scope is intentional but partial. The most natural
extension of this work is to apply the same benchmark-generation
paradigm --- LLM-drafted questions under a structured-output schema,
deterministic citation matching against source text, adversarial slice
scored on grounded refusal, per-level anchored LLM evaluator --- to
multi-paper QA, where the system must gather evidence across multiple
documents to answer a single question. The row's schema structure would
generalize to \emph{(paper\_id, section\_label, alternatives)} triples,
and the matcher already operates per-paper. We expect to release a
multi-paper companion benchmark in a follow-up.

\subsection{Acknowledgements}\label{acknowledgements}

We thank Chris Lengerich for his advice on the structure and content of
this paper.

\subsection{References}\label{references}

\begin{itemize}
\tightlist
\item
  Dasigi, P., Lo, K., Beltagy, I., Cohan, A., Smith, N. A., \& Gardner,
  M. (2021). \href{https://aclanthology.org/2021.naacl-main.365/}{A
  Dataset of Information-Seeking Questions and Answers Anchored in
  Research Papers}. \emph{NAACL 2021}. DOI:
  \href{https://doi.org/10.18653/v1/2021.naacl-main.365}{10.18653/v1/2021.naacl-main.365}.
\item
  Fan, A., Jernite, Y., Perez, E., Grangier, D., Weston, J., \& Auli, M.
  (2019). \href{https://aclanthology.org/P19-1346/}{ELI5: Long Form
  Question Answering}. \emph{ACL 2019}. DOI:
  \href{https://doi.org/10.18653/v1/p19-1346}{10.18653/v1/P19-1346}.
\item
  Gu, J., Jiang, X., Shi, Z., Tan, H., Zhai, X., Xu, C., et al.~(2025).
  \href{https://doi.org/10.1016/j.xinn.2025.101253}{A Survey on
  LLM-as-a-Judge}. \emph{The Innovation}. DOI:
  \href{https://doi.org/10.1016/j.xinn.2025.101253}{10.1016/j.xinn.2025.101253}.
\item
  Hong, Y., Yao, H., Shen, B., et al.~(2025).
  \href{https://arxiv.org/abs/2601.08654}{RULERS: Locked Rubrics and
  Evidence-Anchored Scoring for Robust LLM Evaluation}. \emph{arXiv
  preprint 2601.08654}.
\item
  Jiang, Z., Sun, M., Liang, L., \& Zhang, Z. (2025).
  \href{https://doi.org/10.1145/3701716.3716889}{Retrieve, Summarize,
  Plan: Advancing Multi-hop Question Answering with an Iterative
  Approach}. \emph{Companion Proceedings of the ACM on Web Conference
  (WWW '25)}. DOI:
  \href{https://doi.org/10.1145/3701716.3716889}{10.1145/3701716.3716889}.
\item
  Liu, N. F., Lin, K., Hewitt, J., Paranjape, A., Bevilacqua, M.,
  Petroni, F., \& Liang, P. (2024).
  \href{https://aclanthology.org/2024.tacl-1.9/}{Lost in the Middle: How
  Language Models Use Long Contexts}. \emph{Transactions of the
  Association for Computational Linguistics (TACL)}. DOI:
  \href{https://doi.org/10.1162/tacl_a_00638}{10.1162/tacl\_a\_00638}.
\item
  Liu, Y., Iter, D., Xu, Y., Wang, S., Xu, R., \& Zhu, C. (2023).
  \href{https://aclanthology.org/2023.emnlp-main.153/}{G-Eval: NLG
  Evaluation using GPT-4 with Better Human Alignment}. \emph{EMNLP
  2023}. DOI:
  \href{https://doi.org/10.18653/v1/2023.emnlp-main.153}{10.18653/v1/2023.emnlp-main.153}.
\item
  Lu, P., Mishra, S., Xia, T., Qiu, L., Chang, K.-W., Zhu, S.-C.,
  Tafjord, O., Clark, P., \& Kalyan, A. (2022).
  \href{https://proceedings.neurips.cc/paper_files/paper/2022/hash/11332b6b6cf4485b84afadb1352d3a9a-Abstract-Conference.html}{Learn
  to Explain: Multimodal Reasoning via Thought Chains for Science
  Question Answering}. \emph{NeurIPS 2022}. DOI:
  \href{https://doi.org/10.52202/068431-0182}{10.52202/068431-0182}.
\item
  Ma, J. J., Hashemi, M., Yazdanbakhsh, A., Swersky, K., Press, O., Li,
  E., Reddi, V. J., \& Ranganathan, P. (2025).
  \href{https://arxiv.org/abs/2511.06090}{SWE-fficiency: Can Language
  Models Optimize Real-World Repositories on Real Workloads?}.
  \emph{arXiv preprint 2511.06090}.
\item
  Mazeika, M., Gatti, A., Menghini, C., et al.~(2025).
  \href{https://arxiv.org/abs/2510.26787}{Remote Labor Index: Measuring
  AI Automation of Remote Work}. \emph{arXiv preprint 2510.26787}.
\item
  Menick, J., Trebacz, M., Mikulik, V., Aslanides, J., Song, F.,
  Chadwick, M., Glaese, M., Young, S., Campbell-Gillingham, L., Irving,
  G., \& McAleese, N. (2022).
  \href{https://arxiv.org/abs/2203.11147}{Teaching language models to
  support answers with verified quotes}. \emph{arXiv preprint
  2203.11147}. DOI:
  \href{https://doi.org/10.48550/arXiv.2203.11147}{10.48550/arXiv.2203.11147}.
\item
  Nair, I., Somasundaran, S., Saha, A., \& Bansal, M. (2023).
  \href{https://aclanthology.org/2023.findings-emnlp.972/}{Drilling Down
  into the Discourse Structure with LLMs for Long Document Question
  Answering}. \emph{Findings of EMNLP 2023}. DOI:
  \href{https://doi.org/10.18653/v1/2023.findings-emnlp.972}{10.18653/v1/2023.findings-emnlp.972}.
\item
  Patwardhan, T., et al.~(2025).
  \href{https://arxiv.org/abs/2510.04374}{GDPval: Evaluating AI Model
  Performance on Real-World Economically Valuable Tasks}. \emph{arXiv
  preprint 2510.04374}. DOI:
  \href{https://doi.org/10.70777/si.v2i4.17197}{10.70777/si.v2i4.17197}.
\item
  Song, M., Zheng, M., \& Xu, C. (2026).
  \href{https://arxiv.org/abs/2603.11027}{Beyond the Illusion of
  Consensus: From Surface Heuristics to Knowledge-Grounded Evaluation in
  LLM-as-a-Judge}. \emph{arXiv preprint 2603.11027}.
\item
  Vidgen, B., Fennelly, A., et al.~(2025).
  \href{https://arxiv.org/abs/2509.25721}{The AI Productivity Index:
  APEX-v1-extended}. \emph{arXiv preprint 2509.25721}.
\item
  Wang, M., Chen, L., Cheng, F., Liao, S., Zhang, X., Wu, B., Yu, H.,
  Xu, N., Zhang, L., Luo, R., Li, Y., Yang, M., Huang, F., \& Li, Y.
  (2024a). \href{https://aclanthology.org/2024.emnlp-main.322/}{Leave No
  Document Behind: Benchmarking Long-Context LLMs with Extended
  Multi-Doc QA}. \emph{EMNLP 2024}. DOI:
  \href{https://doi.org/10.18653/v1/2024.emnlp-main.322}{10.18653/v1/2024.emnlp-main.322}.
\item
  Wang, Y., Lipka, N., Rossi, R. A., Siu, A., Zhang, R., \& Derr, T.
  (2024b). \href{https://doi.org/10.1609/aaai.v38i17.29889}{Knowledge
  Graph Prompting for Multi-Document Question Answering}. \emph{AAAI
  2024}. DOI:
  \href{https://doi.org/10.1609/aaai.v38i17.29889}{10.1609/aaai.v38i17.29889}.
\end{itemize}

\subsection{Appendix A: Alternatives
considered}\label{appendix-a-alternatives-considered}

This appendix records the design alternatives we considered and
rejected, organized by which piece of the benchmark they would have
replaced. The intent is to make the reasoning behind the current design
auditable: every choice in the main text has at least one credible
alternative, and most of the rejections were observed empirically rather
than chosen on first principles.

\subsubsection{Schema alternatives}\label{schema-alternatives}

The first iteration of \texttt{expected\_references} was a flat
\texttt{list{[}str{]}} of supporting passages. It worked for
single-section lookup questions but could not express ``the answer must
touch sections A and B'' without conflating the two requirements into a
single bag of strings; we replaced it with the current
\texttt{SectionEvidence} structure when we added multi-hop questions and
discovered that the matcher had no way to distinguish ``missed an entire
required section'' from ``cited the right section but a different valid
alternative within it.'' A single-canonical-reference design --- closer
to QASPER and easier to author --- was rejected for the opposite reason:
we observed the model regularly citing thorough, grounded passages that
differed from the dataset's chosen one, and a single-canonical schema
would have scored those as wrong. Free-text grounding (no structured
references at all, just an LLM evaluator looking at the prose) was
rejected because the evaluator is not a reliable groundedness signal
without the source paper in its prompt --- which we measured directly
when our LLM-evaluated \texttt{groundedness} score collapsed to 5.000
across all systems.

\subsubsection{Metric alternatives}\label{metric-alternatives}

The original metric set included \texttt{citation\_recall} (the fraction
of expected references the model matched), which we replaced with
\texttt{section\_coverage} because recall conflates ``missed a required
hop'' with ``cited one of three valid alternatives within a hop'' ---
both produce a less-than-perfect recall, but they describe completely
different failure modes. Section coverage is structurally aware: the
AND-across-sections, OR-within-alternatives semantics maps directly to
the question structure rather than counting matches in a flat list. The
original adversarial metric was strict in a different way: it required
the model to cite \emph{only} passages from a pre-listed refutation set,
which severely under-scored correct refusals that cited additional
grounded supporting evidence the dataset could not exhaustively
pre-enumerate. Replacing it with ``all cited passages are grounded in
the paper text'' (using the same matcher as \texttt{citation\_accuracy})
lifted refusal correctness by roughly 65 points across providers without
changing model behavior --- the entire gap was in the metric, not in the
systems being measured. We also removed computed \texttt{groundedness}
from the evaluator dimensions for the reason described above; a
paper-text-augmented evaluator could in principle outperform the
deterministic matcher on paraphrased citations, but at significant token
cost and with weaker reproducibility guarantees, and we preferred to
delegate groundedness to the matcher. Finally, we considered a
semantic-similarity citation matcher (embedding cosine over the model's
quote and the paper text) and rejected it for the
\texttt{citation\_accuracy} metric specifically: we want to fail
fabrication, and an embedding match between a fabricated-but-on-topic
quote and the paper would pass. The harness-side citation contract
demands verbatim quotation; the matcher should enforce it.

\subsubsection{Evaluator alternatives}\label{evaluator-alternatives}

The current evaluator is a single Gemini model grading all systems
including itself, which is the cheapest configuration and the one most
vulnerable to same-family bias. A cross-family evaluator (Claude grading
Gemini and OpenAI, Gemini grading Claude, and so on) would mitigate that
bias but multiplies grading cost and adds a normalization step across
evaluators with different scoring tendencies; we have flagged this as an
open item for the next release. A multi-evaluator consensus (mean of
three different evaluators per row) would give a stronger signal still,
at roughly 3× cost and with harder-to-debug score disputes. A more
interesting alternative --- and a candidate for a v2 evaluator if
\texttt{factual\_accuracy} continues to saturate as we add weaker base
models --- is enumeration-then-score judging: a two-pass design in which
the evaluator first enumerates each factual claim in the actual answer
and marks each verified, unverified, or contradicted, and the 1--5 score
is derived deterministically from those counts. This would remove the
evaluator's subjective number-picking entirely and yield a much more
discriminating signal at the cost of a second LLM call per row. The
fourth alternative considered was paper-text-in-prompt judging: pass the
full paper or a relevant excerpt to the evaluator so it can verify
groundedness independently of the matcher. We rejected this for cost
(papers run to 200 KB of text, which multiplies across rows and models)
and because LLM evaluators tend to over-credit plausible-sounding
paraphrases even when given the source text --- the matcher is a
stricter and cheaper check.

\subsubsection{Question-generation
alternatives}\label{question-generation-alternatives}

The gold standard for question quality is human authorship, which we
deferred for cost reasons (we originally budgeted for graduate-student
labelers; we chose to skip the labeled subset in v1 and accept the
resulting limitation). The current pipeline uses Gemini 3.1 Pro to draft
and augment questions under a structured-output schema, with the
citation matcher and the LLM evaluator surfacing mistakes during
benchmarking --- a compromise that gives us scale without human
validation but ties the dataset's distribution to one model family's
question-writing style. The original pipeline used a single-pass
generation step that produced refutation sets too thin (often a single
alternative, or none) for the adversarial metric to work. The current
pipeline folds an augmentation step into the default generation flow,
widening each \texttt{SectionEvidence} with additional valid
alternatives from the same section. That widening is what made the new
adversarial scoring viable --- without it, the metric would have
continued to penalize correct refusals for citing
legitimate-but-unenumerated supporting passages.

\subsubsection{Construction tradeoffs}\label{construction-tradeoffs}

Three deliberate decisions are worth recording, since each shapes how
the results should be read.

\begin{itemize}
\tightlist
\item
  \textbf{No human annotators in v1.} We considered working with a
  cohort of graduate-student labelers to validate a subset. We deferred
  this: the LLM-generated dataset plus the deterministic citation
  matcher gives us a usable signal without it, thought the absence of
  human labels is recorded explicitly as a limitation. A future
  labeler-augmented subset would let us report inter-annotator agreement
  and calibrate the LLM evaluator against human scores.
\item
  \textbf{Public release on HuggingFace, not private hold-out.} We
  discussed whether to keep the dataset private to minimize test leakage
  into future model training. We chose public release
  (\href{https://huggingface.co/datasets/khoj-ai/ResearchQA}{\texttt{khoj-ai/ResearchQA}})
  to make the benchmark useful to the community; we accept that future
  model versions may have seen some of these papers and questions during
  training, and treat that as an unavoidable cost of an open benchmark.
\item
  \textbf{No reference-based n-gram metrics (ROUGE, BERTScore).} We
  considered using these measurements as calibration alongside the LLM
  evaluator. We dropped them --- they reward surface overlap with the
  expected answer rather than the citation-grounded faithfulness we
  actually care about, and adding them would have diluted the signal of
  the metrics we kept.
\end{itemize}

\subsubsection{Harness-vs-baseline
comparison}\label{harness-vs-baseline-comparison}

The harness ships a \texttt{-\/-baseline} mode that bypasses the
OpenPaper retrieval and citation contract entirely, sending the question
and the raw PDF directly to the LLM. In principle, the side-by-side
comparison of harness mode against baseline mode for the same model
isolates the contribution of the harness from the contribution of the
underlying LLM, and was originally intended to be the central comparison
the benchmark answers. We ran early pilots in baseline mode and observed
that on the answer-quality metrics (\texttt{factual\_accuracy},
\texttt{completeness}) the gap between harness and baseline was small
for frontier models --- the underlying LLM was already strong enough
that adding retrieval and citation-contract scaffolding did not
meaningfully change response correctness. Given that the deterministic
citation metrics by design cannot be evaluated in baseline mode
(baseline outputs are free-form prose without a structured citation
contract), and that running the full eight-model baseline sweep would
roughly double per-row inference and grading cost without producing
comparable citation-grounding signal, we elected to omit the baseline
comparison from the headline results. The mode remains in the harness
for future use; a more interesting comparison once we add weaker base
models, or once we want to specifically argue for the value of the
retrieval layer on long-context-limited models, is to selectively re-run
baseline on a subset of providers where the harness contribution is most
likely to bind.

\begin{center}\rule{0.5\linewidth}{0.5pt}\end{center}

\subsection{Appendix B: Code}\label{appendix-b-code}

The dataset's structured-output schema is defined as a hierarchy of
\href{https://docs.pydantic.dev/}{Pydantic} models in
\texttt{evals/generate\_dataset.py}. The top-level
\texttt{PaperEvalGeneration} is what Gemini 3.1 Pro is asked to produce
per paper; the \texttt{Field} descriptions double as part of the prompt,
since the structured-output API surfaces them to the model. The full
hierarchy is reproduced below in dependency order.

Three sparse per-type fields carry additional grading context:
\texttt{metadata\_required\_sections} annotates the integration target
on multi-hop rows, \texttt{metadata\_false\_premise} and
\texttt{expected\_refusal} annotate the failure mode on adversarial
rows, and \texttt{judge\_rubric} carries a per-question rubric that the
LLM evaluator consults when scoring comprehension answers.

\begin{Shaded}
\begin{Highlighting}[]
\KeywordTok{class}\NormalTok{ SectionEvidence(BaseModel):}
    \CommentTok{"""Evidence from one section/area of a paper.}

\CommentTok{    A question may require evidence from multiple sections (multi{-}hop) or just}
\CommentTok{    one (lookup/comprehension). Within a section, the alternatives list holds}
\CommentTok{    multiple verbatim quotes that each independently support the answer — the}
\CommentTok{    model only needs to cite ONE of them to satisfy that section.}
\CommentTok{    """}

\NormalTok{    section\_label: }\BuiltInTok{str} \OperatorTok{=}\NormalTok{ Field(}
\NormalTok{        description}\OperatorTok{=}\StringTok{"The section/area the evidence comes from, e.g. \textquotesingle{}Methods\textquotesingle{}, "}
        \StringTok{"\textquotesingle{}Results\textquotesingle{}, \textquotesingle{}Table 3\textquotesingle{}, \textquotesingle{}Discussion\textquotesingle{}"}
\NormalTok{    )}
\NormalTok{    alternatives: }\BuiltInTok{list}\NormalTok{[}\BuiltInTok{str}\NormalTok{] }\OperatorTok{=}\NormalTok{ Field(}
\NormalTok{        description}\OperatorTok{=}\NormalTok{(}
            \StringTok{"2{-}3 verbatim quotes from this section that EACH INDEPENDENTLY support "}
            \StringTok{"the answer. Each must be character{-}for{-}character from the paper, "}
            \StringTok{"50{-}200 words, and substantively distinct (different passages, not "}
            \StringTok{"paraphrases of each other). If only one valid passage exists, return "}
            \StringTok{"a single{-}element list — never fabricate."}
\NormalTok{        )}
\NormalTok{    )}


\KeywordTok{class}\NormalTok{ LookupQuestion(BaseModel):}
\NormalTok{    question: }\BuiltInTok{str} \OperatorTok{=}\NormalTok{ Field(}
\NormalTok{        description}\OperatorTok{=}\StringTok{"A factual question answerable by finding an exact passage "}
        \StringTok{"in the paper"}
\NormalTok{    )}
\NormalTok{    expected\_answer: }\BuiltInTok{str} \OperatorTok{=}\NormalTok{ Field(}
\NormalTok{        description}\OperatorTok{=}\StringTok{"The correct answer, based on the paper\textquotesingle{}s content"}
\NormalTok{    )}
\NormalTok{    expected\_references: }\BuiltInTok{list}\NormalTok{[SectionEvidence] }\OperatorTok{=}\NormalTok{ Field(}
\NormalTok{        description}\OperatorTok{=}\NormalTok{(}
            \StringTok{"Exactly ONE SectionEvidence covering the section that contains the "}
            \StringTok{"answer, with 2{-}3 alternative verbatim quotes inside."}
\NormalTok{        )}
\NormalTok{    )}


\KeywordTok{class}\NormalTok{ ComprehensionQuestion(BaseModel):}
\NormalTok{    question: }\BuiltInTok{str} \OperatorTok{=}\NormalTok{ Field(}
\NormalTok{        description}\OperatorTok{=}\StringTok{"An abstractive question about themes, methodology, "}
        \StringTok{"implications, or critique"}
\NormalTok{    )}
\NormalTok{    expected\_answer: }\BuiltInTok{str} \OperatorTok{=}\NormalTok{ Field(}
\NormalTok{        description}\OperatorTok{=}\StringTok{"A well{-}reasoned answer drawing on the paper\textquotesingle{}s content"}
\NormalTok{    )}
\NormalTok{    expected\_references: }\BuiltInTok{list}\NormalTok{[SectionEvidence] }\OperatorTok{=}\NormalTok{ Field(}
\NormalTok{        description}\OperatorTok{=}\NormalTok{(}
            \StringTok{"1{-}2 SectionEvidence objects covering the section(s) that support the "}
            \StringTok{"answer. Each contains 2{-}3 alternative verbatim quotes."}
\NormalTok{        )}
\NormalTok{    )}
\NormalTok{    judge\_rubric: }\BuiltInTok{str} \OperatorTok{=}\NormalTok{ Field(}
\NormalTok{        description}\OperatorTok{=}\StringTok{"3{-}5 evaluation criteria for an LLM judge to score answers "}
        \StringTok{"on a 1{-}5 scale"}
\NormalTok{    )}


\KeywordTok{class}\NormalTok{ MultiHopQuestion(BaseModel):}
\NormalTok{    question: }\BuiltInTok{str} \OperatorTok{=}\NormalTok{ Field(}
\NormalTok{        description}\OperatorTok{=}\NormalTok{(}
            \StringTok{"A question that REQUIRES synthesizing information from \textgreater{}=2 distant "}
            \StringTok{"sections of the paper. The answer must NOT be obtainable from any "}
            \StringTok{"single passage. Examples: comparing a result to a stated baseline; "}
            \StringTok{"checking whether a discussion caveat invalidates a headline claim; "}
            \StringTok{"computing a derived quantity from numbers spread across methods + "}
            \StringTok{"results + tables."}
\NormalTok{        )}
\NormalTok{    )}
\NormalTok{    expected\_answer: }\BuiltInTok{str} \OperatorTok{=}\NormalTok{ Field(}
\NormalTok{        description}\OperatorTok{=}\StringTok{"Reasoned answer that explicitly combines facts from each "}
        \StringTok{"required section."}
\NormalTok{    )}
\NormalTok{    expected\_references: }\BuiltInTok{list}\NormalTok{[SectionEvidence] }\OperatorTok{=}\NormalTok{ Field(}
\NormalTok{        description}\OperatorTok{=}\NormalTok{(}
            \StringTok{"TWO OR MORE SectionEvidence objects, one per required section. The "}
            \StringTok{"answer must combine evidence from all of them. Within each "}
            \StringTok{"SectionEvidence, provide 2{-}3 alternative verbatim quotes from that "}
            \StringTok{"section — the model only needs to cite ONE alternative per section "}
            \StringTok{"to satisfy that hop."}
\NormalTok{        )}
\NormalTok{    )}
\NormalTok{    reasoning\_chain: }\BuiltInTok{str} \OperatorTok{=}\NormalTok{ Field(}
\NormalTok{        description}\OperatorTok{=}\NormalTok{(}
            \StringTok{"One{-}sentence description of how the hops connect, e.g. \textquotesingle{}Methods "}
            \StringTok{"reports n=240; Table 3 reports effect=0.4; Discussion notes "}
            \StringTok{"uncontrolled confounder X — combine to assess effective power.\textquotesingle{}"}
\NormalTok{        )}
\NormalTok{    )}
\NormalTok{    judge\_rubric: }\BuiltInTok{str} \OperatorTok{=}\NormalTok{ Field(}
\NormalTok{        description}\OperatorTok{=}\NormalTok{(}
            \StringTok{"3{-}5 evaluation criteria for an LLM judge (1{-}5 scale). MUST include "}
            \StringTok{"an explicit criterion: \textquotesingle{}answer integrates information from each "}
            \StringTok{"required section.\textquotesingle{}"}
\NormalTok{        )}
\NormalTok{    )}


\KeywordTok{class}\NormalTok{ AdversarialQuestion(BaseModel):}
\NormalTok{    question: }\BuiltInTok{str} \OperatorTok{=}\NormalTok{ Field(}
\NormalTok{        description}\OperatorTok{=}\NormalTok{(}
            \StringTok{"A question with a FALSE PREMISE about the paper, or asking about "}
            \StringTok{"something the paper does NOT address. The question must sound "}
            \StringTok{"plausible — a careless reader would not notice the issue. Examples: "}
            \StringTok{"ask about a placebo arm when the design is single{-}arm; ask why "}
            \StringTok{"authors used Bayesian methods when they used frequentist; ask about "}
            \StringTok{"a population the paper explicitly excludes; ask about a metric the "}
            \StringTok{"paper did not report."}
\NormalTok{        )}
\NormalTok{    )}
\NormalTok{    false\_premise: }\BuiltInTok{str} \OperatorTok{=}\NormalTok{ Field(}
\NormalTok{        description}\OperatorTok{=}\NormalTok{(}
            \StringTok{"One sentence stating exactly what is wrong with the question, e.g. "}
            \StringTok{"\textquotesingle{}The study has no placebo arm — it is single{-}arm with historical "}
            \StringTok{"controls.\textquotesingle{}"}
\NormalTok{        )}
\NormalTok{    )}
\NormalTok{    expected\_answer: }\BuiltInTok{str} \OperatorTok{=}\NormalTok{ Field(}
\NormalTok{        description}\OperatorTok{=}\NormalTok{(}
            \StringTok{"The correct response: identify the false premise and/or state that "}
            \StringTok{"the paper does not address the topic. May briefly cite what the "}
            \StringTok{"paper actually says if it directly refutes the premise."}
\NormalTok{        )}
\NormalTok{    )}
\NormalTok{    expected\_refusal: }\BuiltInTok{bool} \OperatorTok{=}\NormalTok{ Field(}
\NormalTok{        description}\OperatorTok{=}\NormalTok{(}
            \StringTok{"True when the correct behavior is to refuse to fabricate / decline "}
            \StringTok{"to answer. Set False only if the paper directly addresses the "}
            \StringTok{"question in an unexpected way (rare). Always provide a value."}
\NormalTok{        ),}
\NormalTok{    )}
\NormalTok{    expected\_references: }\BuiltInTok{list}\NormalTok{[SectionEvidence] }\OperatorTok{=}\NormalTok{ Field(}
\NormalTok{        description}\OperatorTok{=}\NormalTok{(}
            \StringTok{"Usually an empty list. If the paper has a passage that DIRECTLY "}
            \StringTok{"refutes the false premise, include exactly ONE SectionEvidence "}
            \StringTok{"containing 1{-}2 verbatim alternative quotes the model should cite "}
            \StringTok{"when refusing. Otherwise return []."}
\NormalTok{        ),}
\NormalTok{    )}
\NormalTok{    judge\_rubric: }\BuiltInTok{str} \OperatorTok{=}\NormalTok{ Field(}
\NormalTok{        description}\OperatorTok{=}\NormalTok{(}
            \StringTok{"3{-}5 evaluation criteria. MUST include: \textquotesingle{}answer correctly identifies "}
            \StringTok{"the false premise / refuses to fabricate\textquotesingle{} and \textquotesingle{}no fabricated facts "}
            \StringTok{"about topics the paper does not address.\textquotesingle{}"}
\NormalTok{        )}
\NormalTok{    )}


\KeywordTok{class}\NormalTok{ PaperChunk(BaseModel):}
\NormalTok{    section: }\BuiltInTok{str} \OperatorTok{=}\NormalTok{ Field(}
\NormalTok{        description}\OperatorTok{=}\StringTok{"Paper section this chunk comes from, e.g. \textquotesingle{}Results\textquotesingle{}, "}
        \StringTok{"\textquotesingle{}Methods\textquotesingle{}, \textquotesingle{}Discussion\textquotesingle{}"}
\NormalTok{    )}
\NormalTok{    page\_hint: Optional[}\BuiltInTok{int}\NormalTok{] }\OperatorTok{=}\NormalTok{ Field(}
\NormalTok{        default}\OperatorTok{=}\VariableTok{None}\NormalTok{,}
\NormalTok{        description}\OperatorTok{=}\StringTok{"Approximate page number where the chunk appears"}\NormalTok{,}
\NormalTok{    )}
\NormalTok{    description: }\BuiltInTok{str} \OperatorTok{=}\NormalTok{ Field(description}\OperatorTok{=}\StringTok{"Brief description of what this chunk covers"}\NormalTok{)}
\NormalTok{    source\_text: }\BuiltInTok{str} \OperatorTok{=}\NormalTok{ Field(description}\OperatorTok{=}\StringTok{"50{-}300 word excerpt from the paper"}\NormalTok{)}
\NormalTok{    lookup\_question: LookupQuestion}
\NormalTok{    comprehension\_question: ComprehensionQuestion}


\KeywordTok{class}\NormalTok{ PaperEvalGeneration(BaseModel):}
\NormalTok{    paper\_id: }\BuiltInTok{str} \OperatorTok{=}\NormalTok{ Field(description}\OperatorTok{=}\StringTok{"The OpenAlex ID of the paper"}\NormalTok{)}
\NormalTok{    chunks: }\BuiltInTok{list}\NormalTok{[PaperChunk] }\OperatorTok{=}\NormalTok{ Field(}
\NormalTok{        description}\OperatorTok{=}\NormalTok{(}
            \StringTok{"3{-}5 interesting chunks from different sections of the paper. Return "}
            \StringTok{"[] if Section A was not requested in the prompt."}
\NormalTok{        ),}
\NormalTok{    )}
\NormalTok{    multi\_hop\_questions: }\BuiltInTok{list}\NormalTok{[MultiHopQuestion] }\OperatorTok{=}\NormalTok{ Field(}
\NormalTok{        description}\OperatorTok{=}\NormalTok{(}
            \StringTok{"2{-}3 multi{-}hop questions requiring synthesis across distinct, distant "}
            \StringTok{"sections. Return [] if Section B was not requested in the prompt."}
\NormalTok{        ),}
\NormalTok{    )}
\NormalTok{    adversarial\_questions: }\BuiltInTok{list}\NormalTok{[AdversarialQuestion] }\OperatorTok{=}\NormalTok{ Field(}
\NormalTok{        description}\OperatorTok{=}\NormalTok{(}
            \StringTok{"2{-}3 adversarial questions with false premises or asking about topics "}
            \StringTok{"the paper does not address. Return [] if Section C was not requested "}
            \StringTok{"in the prompt."}
\NormalTok{        ),}
\NormalTok{    )}
\end{Highlighting}
\end{Shaded}

The complete benchmark code, including the matcher, the harness, and the
LLM evaluator prompt, is available at the Open Paper repository under
\href{https://github.com/khoj-ai/openpaper/tree/daa2983eb6268fa24043537862688e6b98636459/server/evals}{\texttt{server/evals/}}.

\subsection{Appendix C: Additional
charts}\label{appendix-c-additional-charts}

This appendix collects supplementary figures that illuminate per-domain
and per-question-type behavior beyond what the main results convey.

\pandocbounded{\includegraphics[keepaspectratio,alt={Citation Precision by Domain}]{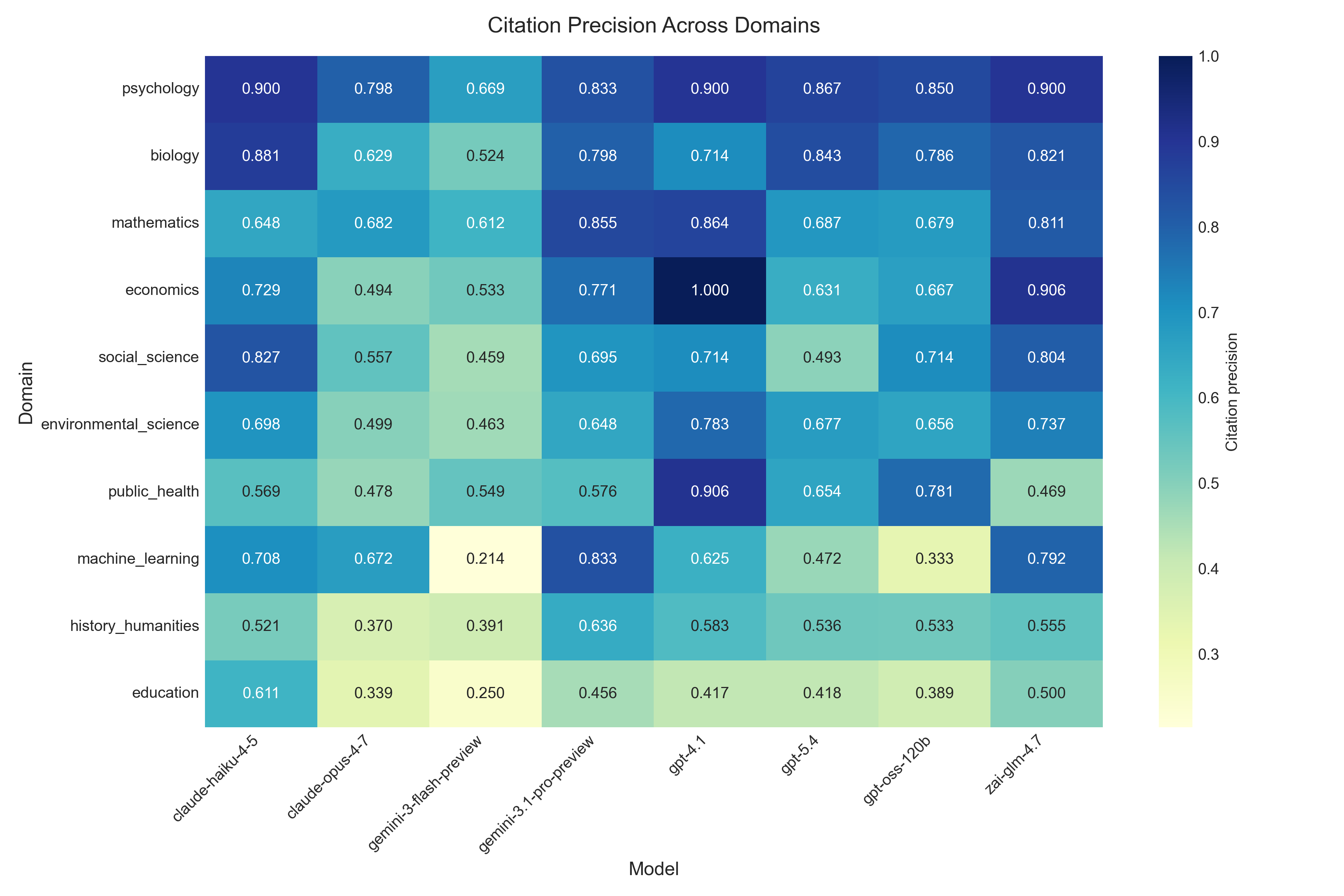}}
\emph{Figure 8: Citation precision by model and domain. The heatmap
exposes domain-by-model interactions invisible in aggregate scores: some
systems are uniformly precise across domains while others swing widely
between best- and worst-domain performance.}

\pandocbounded{\includegraphics[keepaspectratio,alt={Citation Accuracy: Lookup vs Multi-hop}]{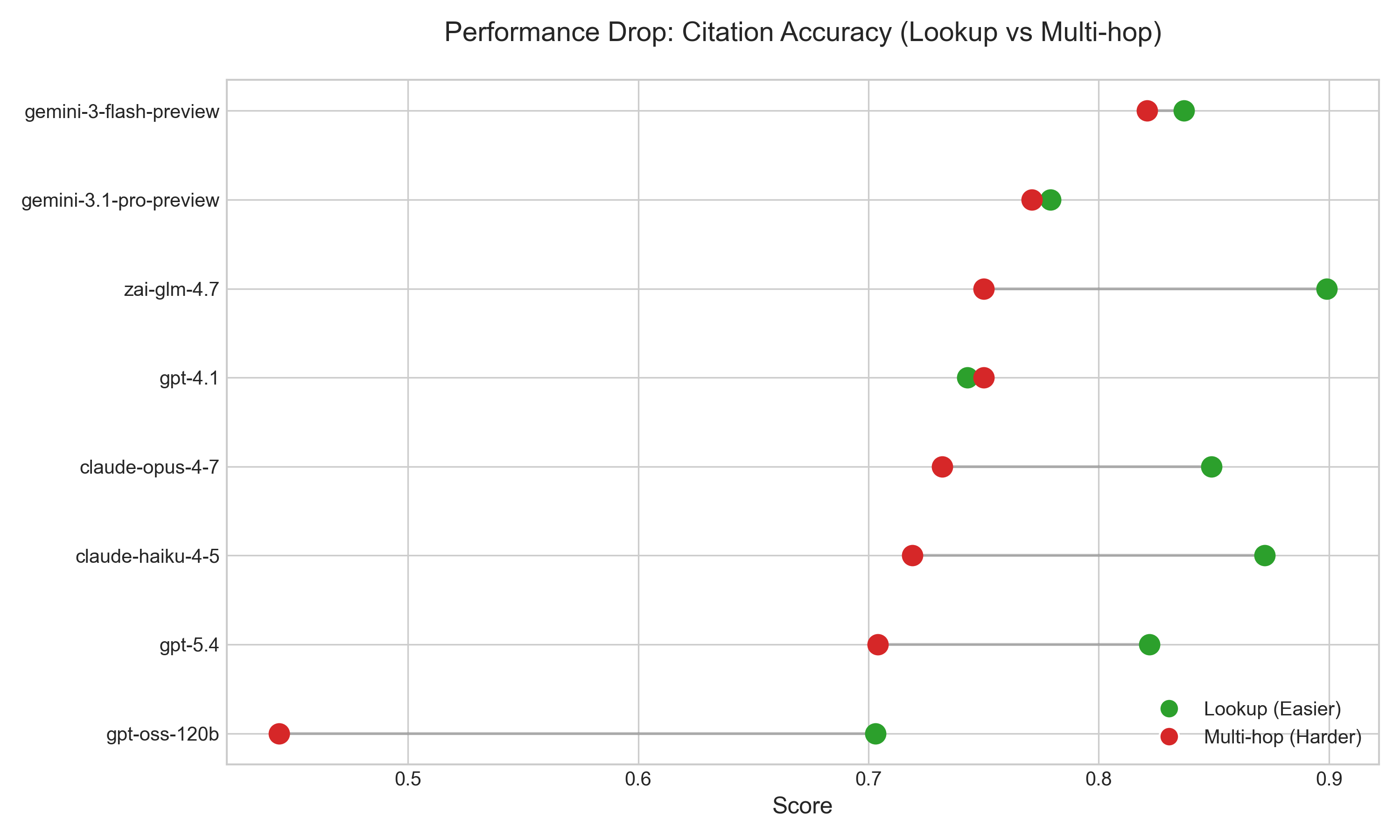}}
\emph{Figure 9: Citation accuracy on lookup versus multi-hop questions.
The companion view to Figure 6: where coverage measures whether the
model touched every required section, accuracy measures whether the
citations the model produced were verifiable substrings of the paper.
Multi-hop accuracy degrades and spreads more than lookup accuracy across
systems.}

\end{document}